\documentclass[letterpaper, 10 pt, conference]{ieeeconf}  

\IEEEoverridecommandlockouts                              

\usepackage{graphics} 
\usepackage{epsfig} 
\usepackage{mathptmx} 
\usepackage{times} 
\usepackage{amsmath} 
\usepackage{bm}
\usepackage{amssymb}  
\usepackage{pgfplots}
\usepackage{graphicx}
\usepackage{subcaption}

\title{\LARGE \bf
 Towards Real-Time Gaussian Splatting: Accelerating 3DGS through Photometric SLAM
}

\author{Yan Song Hu, Dayou Mao, Yuhao Chen, and John Zelek
\thanks{*This work is supported by the Natural Sciences and Engineering Research Council of Canada}
\thanks{All authors are with the Faculty of Systems Design Engineering, University of Waterloo, 200 University Ave W, Waterloo, ON, Canada
        {\tt\small \{y324hu,d6mao,yuhao.chen1, jzelek\}@uwaterloo.ca}}%
}

\begin{document}

\maketitle
\thispagestyle{empty}
\pagestyle{empty}

\begin{abstract}







Initial applications of 3D Gaussian Splatting (3DGS) in Visual Simultaneous Localization and Mapping (VSLAM) demonstrate the generation of high-quality volumetric reconstructions from monocular video streams. However, despite these promising advancements, current 3DGS integrations have reduced tracking performance and lower operating speeds compared to traditional VSLAM. To address these issues, we propose integrating 3DGS with Direct Sparse Odometry, a monocular photometric SLAM system. We have done preliminary experiments showing that using Direct Sparse Odometry point cloud outputs, as opposed to standard structure-from-motion methods, significantly shortens the training time needed to achieve high-quality renders. Reducing 3DGS training time enables the development of 3DGS-integrated SLAM systems that operate in real-time on mobile hardware. These promising initial findings suggest further exploration is warranted in combining traditional VSLAM systems with 3DGS.

\end{abstract}

\section{INTRODUCTION}




Visual Simultaneous Localization and Mapping (VSLAM) is crucial for developing robust mobile robotics. An ideal VSLAM system would reconstruct environments with photorealistic accuracy from live video input. However, traditional VSLAM methods using scene representations such as point clouds and occupancy grids fall short of fully capturing scenes. In contrast, 3D Gaussian Splatting (3DGS) \cite{KKLD23} can generate scenes with enhanced detail and realism. 3DGS is similar to standard triangle rasterization but utilizes 3D Gaussians, which resemble blurry clouds, instead of polygons. Gaussian Splatting works by projecting each Gaussian into the camera, sorting them by depth, and blending them front to back to render the pixels of the rendered image. A representation that is dense, detailed, and can quickly render novel views has many benefits such as enhancing loop closure detection and providing more data for robotic tasks.

\begin{figure}[thpb]
    \centering
    {\renewcommand{\arraystretch}{0}
    \begin{tabular}{c@{}c}
    \begin{subfigure}[b]{.4775\columnwidth}
        \centering
        \includegraphics[width=\columnwidth]{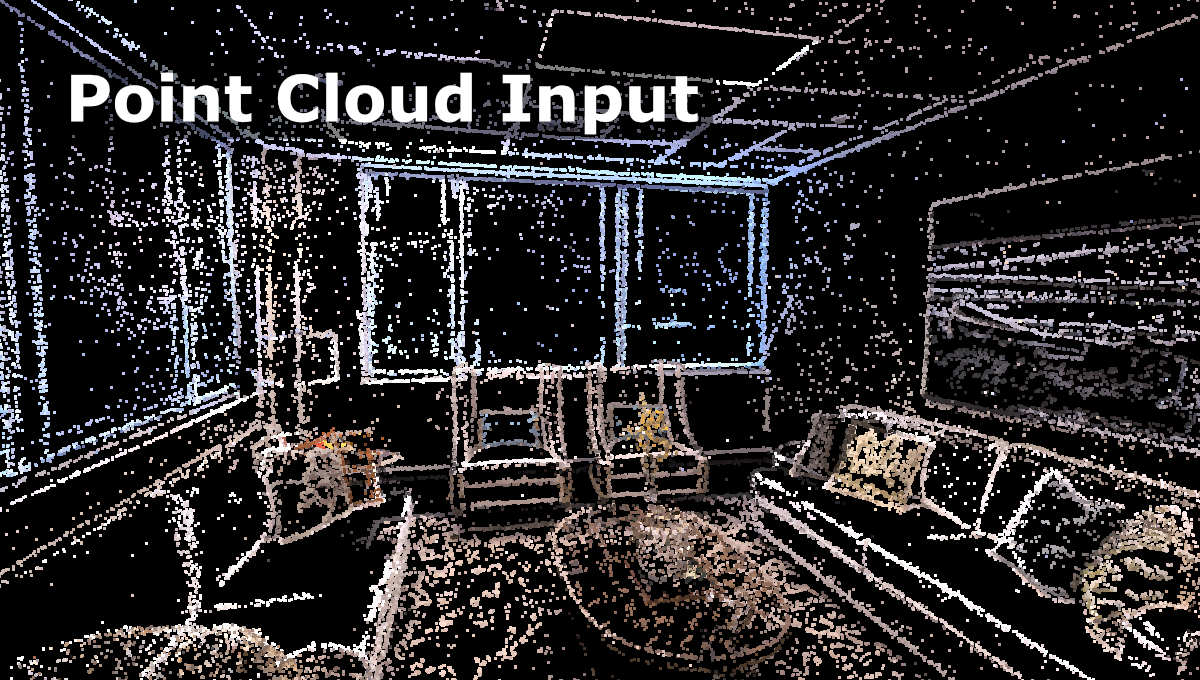}%
    \end{subfigure}&
    \begin{subfigure}[b]{.4775\columnwidth}  
        \centering
        \includegraphics[width=\columnwidth]{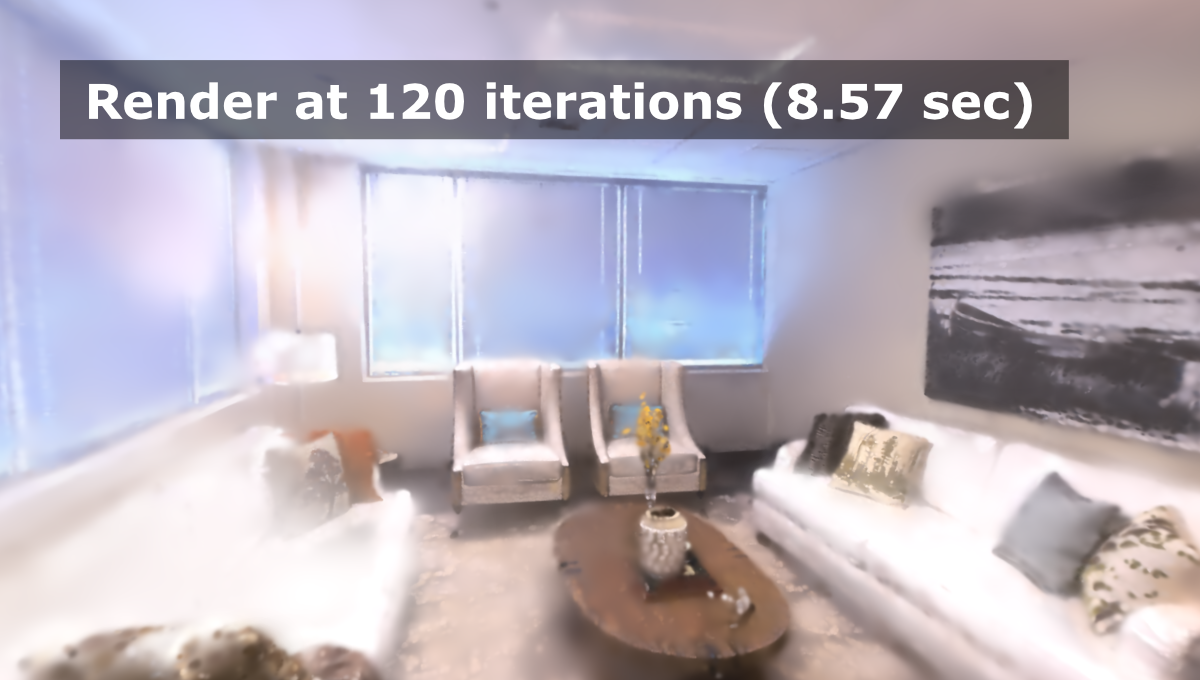}
    \end{subfigure}\\
    \begin{subfigure}[t]{.4775\columnwidth}   
        \centering 
        \includegraphics[width=\textwidth]{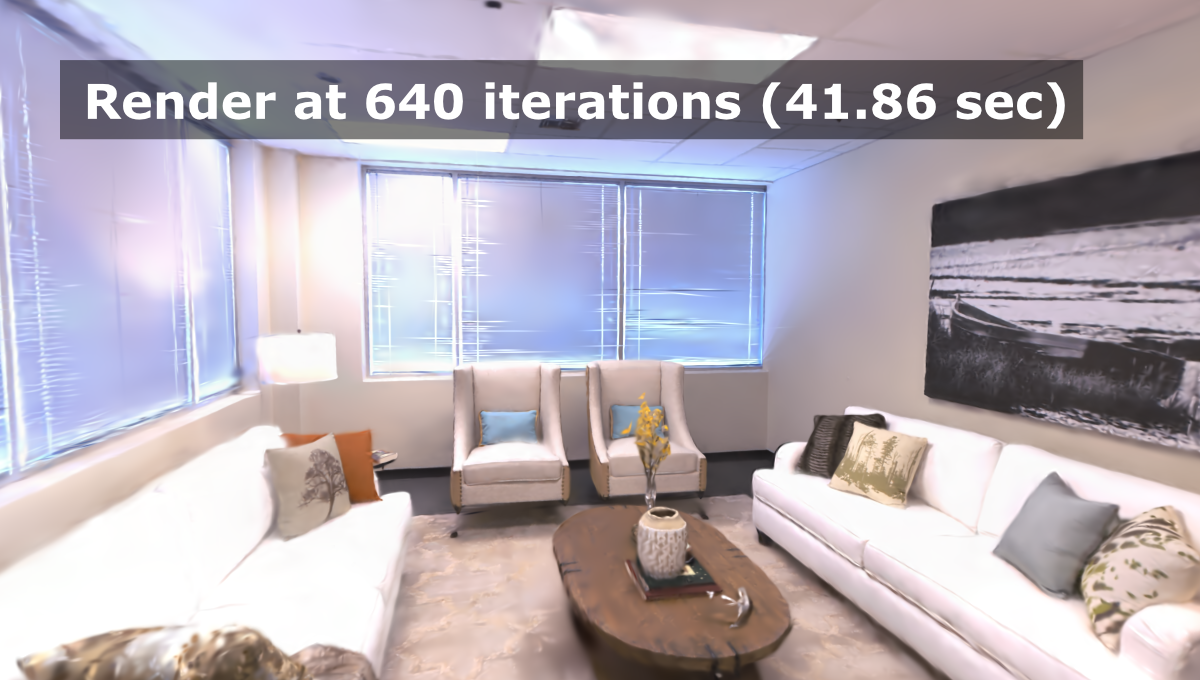}%
    \end{subfigure}&
    \begin{subfigure}[t]{.4775\columnwidth}   
        \centering 
        \includegraphics[width=\columnwidth]{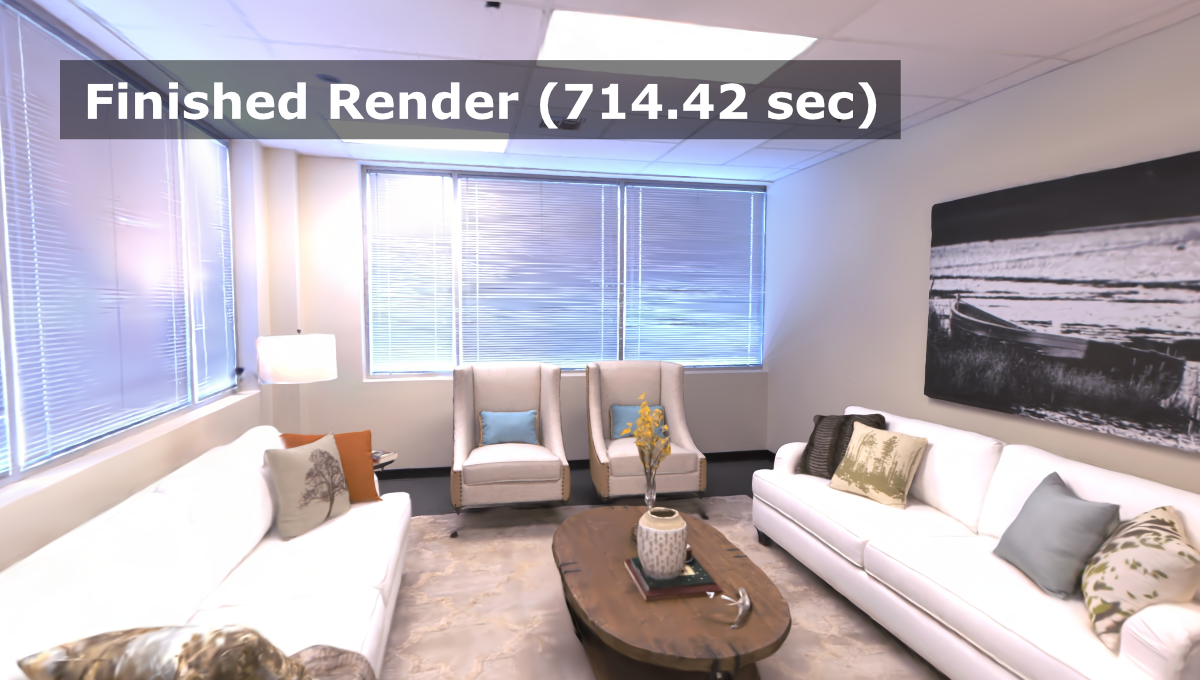}%
    \end{subfigure}
    \end{tabular}}
      \caption{Renders of Replica Room 0 at different training stages. Notice the high quality of renders even at earlier iterations.}
      \label{figurelabel}
   \end{figure}

The first applications of 3DGS for VSLAM such as SplaTAM \cite{keetha2024splatam} were successful in generating 3DGS scenes using live monocular video inputs. However, compared to traditional VSLAM systems like ORBSLAM3 \cite{9440682}, they track less accurately and run slower. The challenges may stem from differences in their tracking methods. Traditional VSLAM calculates poses by tracking feature points across consecutive images. In contrast, 3DGS is incorporated into VSLAM by extending the 3DGS optimization to include poses. The typical 3DGS procedure starts by obtaining a fixed set of poses and initial points from a structure-from-motion system such as Colmap \cite{schoenberger2016sfm}. After converting the initial points into 3D Gaussians, their positions, sizes, and colors are continuously optimized to minimize differences between the rendered and inputted images. If required, Gaussians are added to empty regions or pruned as the optimization progresses. Incorporating VSLAM into 3DGS requires the poses to be optimized simultaneously. Because VSLAM using 3DGS relies on rendered 3D scenes to do tracking, its speed is limited by the need to generate a high-quality 3D scene. For instance, SplaTAM does 40 to 60 training iterations per frame to reach sufficient map quality for accurate tracking \cite{keetha2024splatam}. Therefore, lowering the time it takes to build good maps would improve the speeds of VSLAM systems using 3DGS. 

InstantSplat \cite{fan2024instantsplat} is a technique that can quickly produce high-quality 3DGS maps. It uses DUSt3R \cite{dust3r2023}, a state-of-the-art stereo reconstructor, to generate dense initial point clouds instead of sparse point clouds from typical structure-from-motion. DUSt3R generates points even in regions lacking in features, speeding up the 3DGS process by eliminating the need to create points at those locations.

However, DUSt3R is not the only way to generate high-density point clouds. Our primary contribution is demonstrating that monocular photometric or pixel-based VSLAM systems, such as Direct Sparse Odometry (DSO) \cite{Engel-et-al-pami2018}, can produce high-density point clouds that accelerate 3DGS training. Pixel-based SLAM systems track high-gradient pixels instead of feature points, resulting in denser point clouds due to having more tracking candidates. We further modified DSO to track more pixels not used for pose optimization, increasing point cloud density to levels comparable with DUSt3R. We did experiments showing that inputting modified DSO point clouds and poses into 3DGS instead of using Colmap significantly improves training times, especially early in the training process. Qualitatively good 3DGS renders have been produced in under a minute, as shown in Figure \ref{figurelabel}. Additionally, DSO runs at live speeds, which is faster compared to typical structure-from-motion systems. This contribution is particularly beneficial for VSLAM systems using 3DGS where speed is paramount.

\section{Method}


This section provides a simplified description of DSO and the modifications made. DSO works by tracking a set of pixels across consecutive frames \(i\) and \(j\). It optimizes \(\bm{p}\), the pose, using the photometric loss equation presented below for each pixel in the set:  

\newcommand{\norm}[1]{\left\lVert#1\right\rVert}
\begin{equation*}
E=\norm{
    (\mathit{I}_j[\bm{p}_j]-b_j) -
    \frac{s_j a_j}{s_i a_i}
    (\mathit{I}_i[\bm{p}_i]-b_i)
}
\end{equation*}

Where \(I\) queries the pixel intensity, \(a\) and \(b\) are photometric variables, and \(s\) is the exposure. This function finds the change in pose that best matches the pixel change between frames \(i\) and \(j\) while accounting for lighting changes. 

By tracking high-gradient pixels, DSO can create dense point clouds for its map; however, the point selection is optimized for tracking performance rather than maximizing point density. We found that 3DGS trains faster on denser point clouds than those typically generated for DSO's optimal tracking settings. To enhance 3DGS performance without compromising tracking, we modified DSO to include additional points not used for pose tracking.

The pixel selector of DSO is modified to find uniformly distributed extra pixels in areas lacking tracked pixels. Because these extra pixels can be in difficult to track locations, they only have their position optimized and do not affect the overall pose tracking. Furthermore, some regions in images lack gradients, making tracking impossible. To ensure points exist in these gradient-less areas for 3DGS, we implemented a system that places some points in these regions and sets their positions to the average of nearby tracked points.

\section{Preliminary Results}

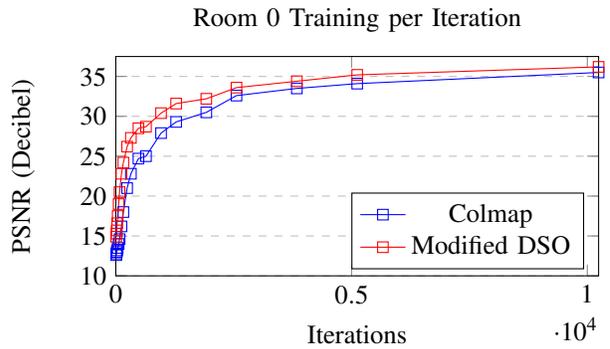
\begin{figure}[htbp]
\begin{tikzpicture}
\begin{axis}[
    title={Room 0 Training per Iteration},
    xlabel={Iterations},
    ylabel={PSNR (Decibel)},
    xmin=0, xmax=10240,
    ymin=10, ymax=37.5,
    xtick={0,5000,10000},
    ytick={0,5,10,15,20,25,30,35},
    legend pos=south east,
    ymajorgrids=true,
    grid style=dashed,
    height=4.5cm,
    width=8cm
]

\addplot[
    color=blue,
    mark=square,
    ]
    coordinates {
    (10,12.6)(15,12.8)(20,13)(30,13.2)(40,13.5)(60,14)(80,14.6)(120,16.2)(160,18)(240,21)(320,22.8)(480,24.7)(640,25)(960,27.9)(1280,29.3)(1920,30.5)(2560,32.6)(3840,33.5)(5120,34.1)(10240,35.5)
    };
\addplot[
    color=red,
    mark=square,
    ]
    coordinates {
    (10,14.9)(15,15.3)(20,15.7)(30,16.6)(40,17.5)(60,19)(80,20.5)(120,22.8)(160,24.2)(240,26.2)(320,27.3)(480,28.5)(640,28.7)(960,30.4)(1280,31.6)(1920,32.2)(2560,33.6)(3840,34.4)(5120,35.2)(10240,36.2)
    };
    \legend{Colmap,Modified DSO};
    
\end{axis}
\end{tikzpicture}
\caption{Peak Signal to Noise Ratio (PSNR) of training of Room 0 of the Replica dataset over time. Higher PSNR is better. Average taken over ten runs.}
\label{graph1}
\end{figure}

Preliminary experiments on the Replica dataset \cite{replica19arxiv} were made to explore the validity of the method. A subset of the results are shown by figure \ref{graph1} and table \ref{table1}.

\begin{table}[htbp]
\caption{Peak Signal to Noise (PNSR) at Specific Iterations}
\begin{center}
\begin{tabular}{|c|c|c|c|c|}
\hline
Iterations & Room0 & Room1 & Room2 & Average \\
\hline
\multicolumn{5}{|c|}{Colmap} \\
\hline
120 & 16.22 & 17.30 & 17.91 & 17.14 \\
\hline
640 & 25.00 & 23.61 & 25.51 & 24.71 \\
\hline
\multicolumn{5}{|c|}{Modified DSO} \\
\hline
120 & 22.81 & 23.28 & 25.61 & 23.90 \\
\hline
640 & 28.74 & 29.90 & 32.04 & 30.23 \\
\hline
\multicolumn{4}{l}{Average of ten runs}
\end{tabular}
\label{table1}
\end{center}
\end{table}

As one can see from the data, using modified DSO point cloud inputs results in the peak signal to noise ratio, which is a measurement of image rendering quality, increasing faster compared to using traditional structure-from-motion point clouds.

\section{CONCLUSIONS}


To summarize, our main contribution is demonstrating that current photometric VSLAM methods can enhance the speed and efficiency of 3DGS. Experimental results show that outputs from photometric VSLAM can accelerate 3DGS training, which leads to faster tracking in VSLAM systems using 3DGS. We hope future work builds on this research by running DSO and 3DGS in parallel, rather than merely using DSO outputs as inputs for 3DGS. However, combining these techniques is not straightforward and requires more research due to their different tracking optimization methods. Despite the difficulties, combining photometric and 3DGS techniques can result in a VSLAM system as fast as the state-of-the-art while providing detailed, dense representations.

\addtolength{\textheight}{-12cm}   


\bibliographystyle{IEEEtran}
\bibliography{Paperbib}

\end{document}